\documentclass[twoside,11pt]{article}

\usepackage{blindtext}

\usepackage{dmlr2e}

\usepackage{times}
\usepackage{latexsym}
\usepackage{graphicx}
\usepackage{booktabs}
\usepackage{amsmath,amsfonts,amssymb}
\usepackage{multirow}

\usepackage{array}
\newcolumntype{C}[1]{>{\centering\arraybackslash}p{#1}}

\usepackage{pifont}
\usepackage[T1]{fontenc}

\usepackage[utf8]{inputenc}

\usepackage{microtype}
\usepackage{wrapfig}

\usepackage{lastpage}
\dmlrheading{24}{2024}{1-\pageref{LastPage}}{2/2/2024; Revised N/A}{N/A}{21-0000}{Leotescu et al} 

\ShortHeadings{Bidirectional Long-Range Parser for Sequential Data Understanding}{DMLR@ICLR-24}
\firstpageno{1}

\begin{document}

\title{Bidirectional Long-Range Parser for Sequential Data Understanding}

%
\author{\name George Leotescu \email leoteg@amazon.com \\
       \addr Amazon Inc.
       \AND
       \name Daniel Voinea \email dvoinea@amazon.com \\
       \addr Amazon Inc.
       \AND
       \name Alin-Ionut Popa \email popaaln@amazon.com \\
       \addr Amazon Inc.}


\maketitle

\begin{abstract}
The transformer is a powerful data modelling framework responsible for remarkable performance on a wide range of tasks. However, they are limited in terms of scalability as it is suboptimal and inefficient to process long-sequence data.
To this purpose we introduce \textbf{BLRP} (\textbf{B}idirectional \textbf{L}ong-\textbf{R}ange \textbf{P}arser), a novel and versatile attention mechanism designed to increase performance and efficiency on long-sequence tasks. It leverages short and long range heuristics in the form of a local sliding window approach combined with a global bidirectional latent space synthesis technique.
We show the benefits and versatility of our approach on vision and language domains by \emph{demonstrating competitive results against state-of-the-art} methods on the Long-Range-Arena and CIFAR benchmarks together with ablations demonstrating the computational efficiency. 
\end{abstract}


\section{Introduction}

\begin{wrapfigure}{r}{0.5\textwidth}
\vspace{-10pt}
\centering
\includegraphics[width=0.5\textwidth]{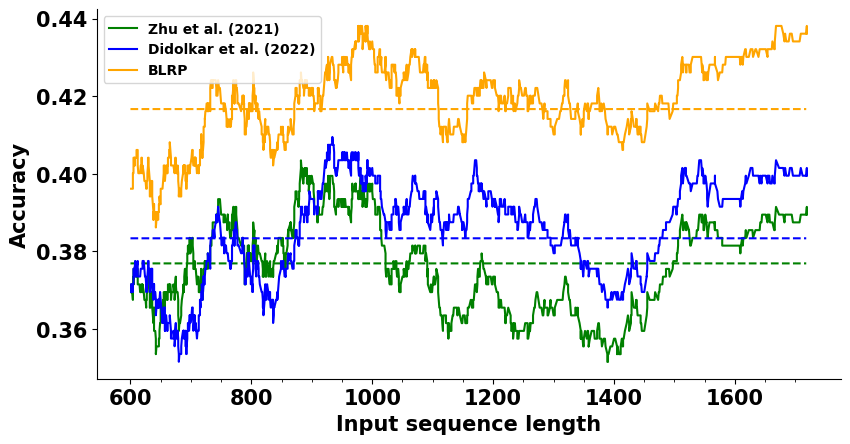}
\vspace{-20pt}
\caption{\footnotesize{\textbf{Performance comparison for different sequence lengths.} We compare our proposed \textbf{BLRP} framework against \cite{didolkartemporal} and \cite{zhu2021long} for different sequence lengths on \texttt{ListOps}. \textbf{BLRP} brings scalable performance gains irrespective to sequence length, showing that the bi-directional mechanism increases the model's representative power on all input length ranges. Dotted lines represent average performance for each method.}}
\vspace{-18pt}
\label{fig:plot_performance}
\end{wrapfigure}

Statistical modeling of sequential data has witnessed increased scientific attention since the appearance of probabilistic learning approaches. Initially, notable approaches such as \cite{norris1998markov,hochreiter1997long,schuster1997bidirectional,cho2014learning,arjovsky2016unitary} applied recurrent processing for this task. Usually they extract discriminative features and capture dominant patterns from the analysed data. Although this procedure has its merits, it is limited in terms of generability. For example, the recurrence aspect encourages sensitivity with respect to the ordering of the elements.
With the emergence of transformer methodology \cite{vaswani_nips_2017} and the introduction of the multi-head attention, the performance boundaries for sequential data understanding tasks were pushed further. It is a powerful mechanism which parses in parallel the sequence elements and performs an implicit embedding statistic, thus emphasizing global relationships among the elements. 

However, this benefit comes at the expense of scalability. 
The attention mechanism, in its primal form, scales quadratically with respect to the sequence size and as a direct consequence it is expensive and suboptimal to parse data structured as long sequences. 

\begin{figure*}[t]
		\includegraphics[width=0.99\linewidth]{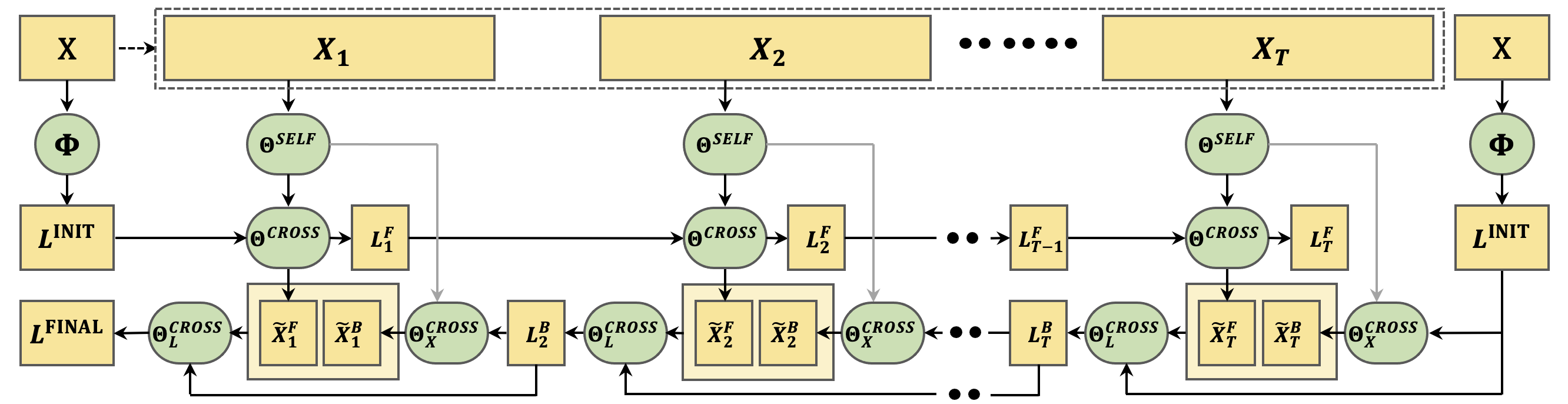}
	\caption{\footnotesize{\textbf{Detailed overview of the proposed \textbf{BLRP} method.} The flow is from left to right for the \emph{forward} pass, followed by right to left for the \emph{backward} pass. The input sequence $\mathbf{X}$ is split into a list of non-overlaping segments $(\mathbf{X}_i)_{i=1}^T$, which is bidirectionally parsed while capturing the overall information into the latent block $\mathbf{L}$, originally initialized with $\mathbf{L}^\mathtt{INIT}$ via function $\Phi$. In turn, $\mathbf{L}^\mathtt{INIT}$ is added at the start of both forward and backward passes and subsequently used as a residual connection. Thus, we obtain the temporal level state representations, $(\mathbf{L}_i^\mathtt{F})_{i=1}^T$ and $(\mathbf{L}_i^\mathtt{B})_{i=T}^1$, which synthesise the aggregated information from the entire sequence inside $\mathbf{L}^\mathtt{FINAL}$. Notice that we optimally aggregate information at spatial level by iteratively conditioning the latent states on the segment embeddings, and at temporal level by utilizing the corresponding forward segment embeddings to update the backward states.}} 
	\label{fig:detailed_overview}
\end{figure*}

To address this limitation, we propose \textbf{B}idirectional \textbf{L}ong-\textbf{R}ange \textbf{P}arser, a novel attention framework designed to efficiently parse long sequences (\textit{i.e.} $> 2,000$) which integrates \emph{(I)} a local-window attention capturing \textit{small scale} correlations between the elements of the sequence and \emph{(II)} a bidirectional aggregation technique which captures recurrently the \textit{large scale} context of the full-sequence into a temporal latent block representation. The intuition behind our approach is to efficiently capture proximal and distant relationships between elements at spatial level (\textit{i.e.} sequence positioning), while taking into account their ordering interpreted as the temporal axis (\textit{i.e.} going forward and backward along the sequence). Thus, we build a data-centric solution to efficiently address the long sequence scalability issue. From the linguistic point of view, our proposed approach is able to recover local discriminative vocabulary items, while successfully grasping the ample context where they are utilized and observe from a bidirectional perspective how they contribute to the textual flow. We experiment on the challenging benchmarks Long-Range-Arena (LRA) \cite{taylong} and CIFAR \cite{krizhevsky2009learning} proving competitive results against state-of-the-art approaches on multiple domains. An overview of our proposed mechanism is illustrated in Figure \ref{fig:detailed_overview}. Our approach is generic, in the sense that it can be applied on top of any sequential data parsing approach from multiple domains (\textit{i.e.} language or vision).

\nocite{stoian2022unstructured}


\section{Related Work}
Transformers \cite{vaswani_nips_2017} made a significant impact across multiple machine learning fields, such as computer vision \cite{carion2020end,radford2021learning}, signal processing \cite{che2021constrained,hu2022transformer,huangencoding} or NLP \cite{layoutlm,xu2020layoutlmv2,tay2021charformer,li2021selfdoc}. They are able to dynamically capture efficient dependencies across the sequence while analysing the entire content in parallel by leveraging the attention mechanism.
While recent research efforts focused on multi-modal attention \cite{li2021selfdoc,appalaraju2021docformer,dosovitskiy2020} and more efficient ways to extract information, the major limitation imposed by quadratic formulation of the attention matrix has been relatively unexplored.
A number of approaches \cite{didolkartemporal,hutchinsblock,zhu2021long,dai2019transformer,rae2019compressive,wu2020lite,zhou2021informer,zhang2021multi,zaheer2020big,beltagy2020longformer} have been proposed to address this problem by focusing on approximating the attention matrix. Our proposed \textbf{BLRP} parses the sequence in a bidirectional fashion by retaining a perceptual latent representation. Moreover, encouraged by the ideas of \cite{zhu2021long} we initialize the latent block representation using a projection of the entire input sequence. 
Another important aspect of our approach is that we are in line with the model requirements and hyperparameter configuration from the LRA benchmark\footnote{\href{https://github.com/google-research/long-range-arena}{https://github.com/google-research/long-range-arena}} to have a fair comparison with current state-of-the-art. 

\nocite{sandu-etal-2022-large}
\nocite{matcovici2023k}
\nocite{krubinski2024watermark}




Our approach is more similar to the work of \cite{zhu2021long}. They consider analysing the sequence in two different ways: \emph{locally}, using a sliding window attention heuristic and \emph{globally}, by dynamically projecting the entire input sequence. The work of \cite{didolkartemporal} is also related to us as they consider splitting the input sequence into chunks and unidirectionaly parsing them into a canonical representation. 

We are aware that there are multiple methods \cite{gu2022train,smith2022simplified,hasani2022liquid,orvieto2023resurrecting} outperforming the LRA benchmark by a large margin  while not applying the above mentioned constraints. However, our goal is to ensure a fair comparison normalized with respect to the total number of the parameters of the model and within the boundaries of the original hyperparameter configuration proposed by the authors of the LRA benchmark. 



\section{Methodology}

In this section we will elaborate the computational details behind our proposed \textbf{BLRP} framework. Let there be an input sequence $\mathbf{X} = (x_1, x_2, \dots, x_N) \in \mathbb{R}^{N \times d}$ containing $N$ elements with dimensionality $d$. 
This is split into a list of $T$ segments $\mathbf{X}_i = (x_{(i - 1) \cdot t + 1}, x_{(i - 1)\cdot t + 2}, \dots, x_{i\cdot t}) \in \mathbb{R}^{t \times d}$, each of equal size $t$. The last segment $\mathbf{X}_T$ is padded to achieve the desired size $t$.
Our model leverages a latent block representation $\mathbf{L} \in \mathbb{R}^{l \times d}$ which is used to retrieve relevant global information. This is achieved via a bidirectional flow over the list of segments $(\mathbf{X}_i)_{i=1}^T$. In essence, $\mathbf{L}$ transitions to a different state as it moves along the bidirectional loop. As a result of the forward and backward pass, we have the following state representations of $\mathbf{L}$: $(\mathbf{L}_i^\mathtt{F})_{i=1}^T$ going forward and $(\mathbf{L}_i^\mathtt{B})_{i=T}^1$ going backward. We retain the final state representation of $\mathbf{L}$ denoted with $\mathbf{L}^\mathtt{FINAL}$ which synthesizes the information from the entire sequence $\mathbf{X}$. For each segment $\mathbf{X}_i$ we include the $1$D positional embeddings following \cite{didolkartemporal}. 



In the following, we will detail the algorithmic steps of our proposed \textbf{BLRP} pipeline.
Firstly, we initialize the latent state using a projection of the raw sequence into the latent space using the function $\Phi : \mathbb{R}^{N \times d} \rightarrow \mathbb{R}^{t \times d}$. Inspired from \cite{zhu2021long} we apply a dynamic projection of the input sequence into the latent space via a learned basis representation. This process is repeated independently for both forward and backward passes. Thus, we have the initial latent state denoted as $\mathbf{L}^\mathtt{INIT} = \Phi(\mathbf{X})$ which is applied at the initial step of each pass, forward and backward. 
The attention operations play a crucial role in our framework. The self-attention operation denoted as $\Theta^\mathtt{SELF}$ has the following formula, $\Theta^\mathtt{SELF}(\mathbf{X}) = softmax(\frac{q^\mathtt{S}(\mathbf{X}){k^\mathtt{S}(\mathbf{X})}^\top}{\sqrt{d}})v^\mathtt{S}(\mathbf{X})$, where functions $k^\mathtt{S}(\cdot)$, $q^\mathtt{S}(\cdot)$ and $v^\mathtt{S}(\cdot)$ represent the keys, queries and values, respectively, together with their embedded projection weights. $\Theta^\mathtt{SELF}$ is used to identify local correlations between the elements within each segment. 

For the forward pass, we process iteratively each input segment $\mathbf{X}_i$ to obtain the latent block states $(\mathbf{L}_i^{\mathtt{F}})_{i=1}^T$  and the forward processed input segment embeddings $(\mathbf{\tilde{X}}_i^{\mathtt{F}})_{i=1}^T$ 

\[
\mathbf{\tilde{X}}_i^{\mathtt{F}}= 
\begin{cases}
	\Theta_X^\mathtt{CROSS}(\Theta^\mathtt{SELF}(\mathbf{X}_i), \mathbf{L}^\mathtt{INIT}),&  i= 1\\
	\Theta_X^\mathtt{CROSS}(\Theta^\mathtt{SELF}(\mathbf{X}_i), \mathbf{L}^\mathtt{F}_{i-1}),              & i > 1
\end{cases}
\]

\[
\mathbf{L}_i^{\mathtt{F}}= 
\begin{cases}
	\Theta_L^\mathtt{CROSS}(\mathbf{L}^\mathtt{INIT}, \mathbf{\tilde{X}}_i^\mathtt{F}),& i= 1\\
	\Theta_L^\mathtt{CROSS}(\mathbf{L}^\mathtt{F}_{i-1}, [\mathbf{\tilde{X}}_i^\mathtt{F} ; \mathbf{L}^\mathtt{INIT}]),              & i > 1
\end{cases}
\]

\noindent where, 

\begin{equation*} 
	\Theta^\mathtt{CROSS}_X(\mathbf{X}, \mathbf{L}) =  softmax(\frac{q^\mathtt{C}_X(\mathbf{X}){k^\mathtt{C}_X(\mathbf{L})}^\top}{\sqrt{d}})v^\mathtt{C}_X(\mathbf{L})
\end{equation*}

\begin{equation*} 
	\Theta^\mathtt{CROSS}_L(\mathbf{X}, \mathbf{L}) =  softmax(\frac{q^\mathtt{C}_L(\mathbf{L}){k^\mathtt{C}_L(\mathbf{X})}^\top}{\sqrt{d}})v^\mathtt{C}_L(\mathbf{X})
\end{equation*}

\begin{wrapfigure}{r}{0.4\textwidth}
\vspace{-10pt}
\centering
\includegraphics[width=0.4\textwidth]{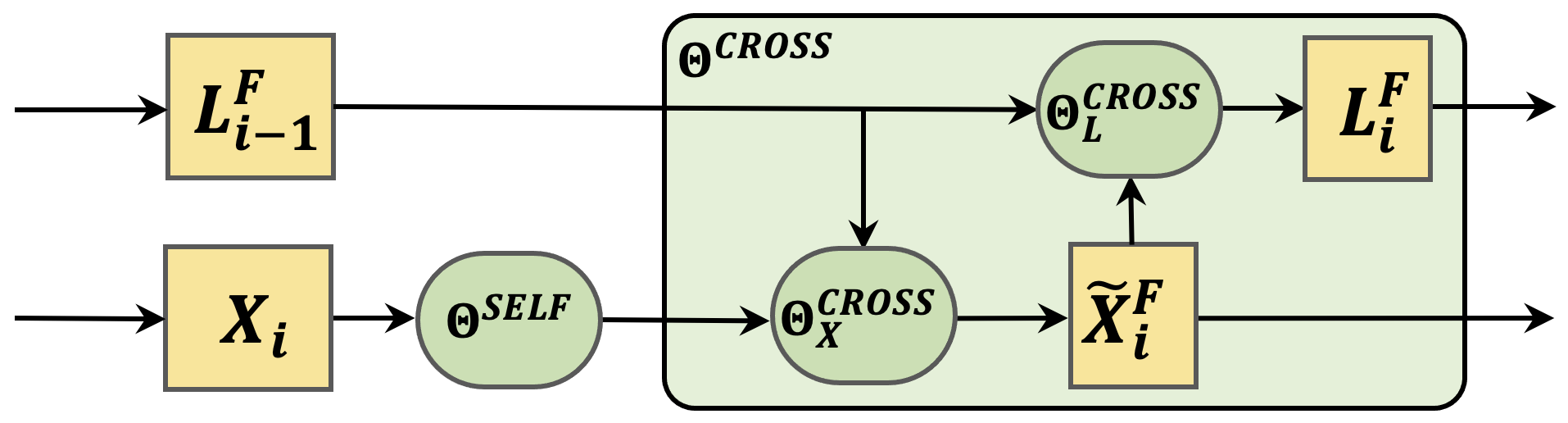}
\vspace{-20pt}
\caption{\footnotesize{\textbf{Module $\Theta^{\mathtt{CROSS}}$ at step \textit{i}.} Forward embeddings $\mathbf{\tilde{X}}_i^{\mathtt{F}}$ and latent state $\mathbf{L}_{i}^\mathtt{F}$ are obtained by an interleave usage of $\Theta_{L}^\mathtt{CROSS}$ and $\Theta_{X}^\mathtt{CROSS}$.}}
\vspace{-18pt}
\label{fig:cross_attention_module}
\end{wrapfigure}

\noindent Functions $k_L^\mathtt{C}(\cdot)$, $q_L^\mathtt{C}(\cdot)$, $v_L^\mathtt{C}(\cdot)$, $k_X^\mathtt{C}(\cdot)$, $q_X^\mathtt{C}(\cdot)$ and $v_X^\mathtt{C}(\cdot)$ are equivalent to $k^\mathtt{S}(\cdot)$, $q^\mathtt{S}(\cdot)$ and $v^\mathtt{S}(\cdot)$. This updating step is illustrated in Figure \ref{fig:cross_attention_module}. In essence, at each time step $i$ we update the segment embeddings, $\mathbf{\tilde{X}}_i^{\mathtt{F}}$, via cross-attention against the current latent block state. Next, we repeat the update process in the same manner to generate the backward states, $(\mathbf{L}_i^{\mathtt{B}})_{i=T}^1$, for the latent block. 
When updating the backward latent states, we let the cross-attention keys attend to the union between the encoded forward segment (\textit{i.e.} $\mathbf{\tilde{X}}_i^{\mathtt{F}}$), the encoded backward segment (\textit{i.e.} $\mathbf{\tilde{X}}_i^{\mathtt{B}}$) and $\mathbf{L}^\mathtt{INIT}$, with

\[
\mathbf{\tilde{X}}_i^{\mathtt{B}}= 
\begin{cases}
	\Theta_X^\mathtt{CROSS}(\Theta^\mathtt{SELF}(\mathbf{X}_i), \mathbf{L}^\mathtt{INIT}),&  i = T\\
	\Theta_X^\mathtt{CROSS}(\Theta^\mathtt{SELF}(\mathbf{X}_i), \mathbf{L}^\mathtt{B}_{i+1}),              & i < T
\end{cases}
\]

\[
\mathbf{L}_i^{\mathtt{B}}= 
\begin{cases}
	\Theta_L^\mathtt{CROSS}(\mathbf{L}^\mathtt{F}_T, [\mathbf{\tilde{X}}_i^\mathtt{F};\mathbf{\tilde{X}}_i^\mathtt{B}]),& i = T \\
	\Theta_L^\mathtt{CROSS}(\mathbf{L}^\mathtt{B}_{i+1}, [\mathbf{\tilde{X}}_i^\mathtt{F};\mathbf{\tilde{X}}_i^\mathtt{B}; \mathbf{L}^{\mathtt{INIT}}]),              & i < T
\end{cases}
\]

Intuitively, this acts as an intertwined spatio-temporal computational grid operating at micro and macro level designed to efficiently extract relevant information encoded within a long sequence. Finally, the last state of the latent block (\textit{i.e.} $\mathbf{L}_1^{\mathtt{B}}$), denoted with $\mathbf{L}^\mathtt{FINAL}$, retrieved after successfully passing the backward pass is incorporated as a bidirectional embedding for the entire sequence $\mathbf{X}$ for other downstream tasks. In our experiments, we couple it as input to an MLP head for a classification task. However, it can be incorporated into any other downstream bottom-up task which involves sequential data understanding (\textit{e.g.} Question Answering, Named-Entity Recognition).

\begin{table}
	\scalebox{0.7}{
		\begin{tabular}{C{33mm}|C{17mm}|C{13mm}|C{21mm}}
			\hline
			\textbf{Model} &  \textbf{\texttt{ListOps}} & \textbf{\texttt{Text}} & \textbf{\texttt{Retrieval}}  \\
			\hline
			\hline

			\cite{wang2020linformer} & $37.38$ & $56.12$ & $79.37$ \\
            \hline
			\cite{xiong2021nystromformer} & $37.34$ & $65.75$ & $81.29$ \\
            \hline
		    \cite{zhu2021long} & $37.5$ & $66.0$ & $81.79$ \\
			\hline
		    \cite{didolkartemporal} & $38.2$ & $82.08$ & $76.91$  \\
			\hline 
			\textbf{BLRP} & $\mathbf{41.43}$ & $\mathbf{82.83}$ & $\mathbf{83.43}$ \\
			\hline
	\end{tabular}}
	\hfill
	\scalebox{0.62}{
	\begin{tabular}{C{40mm}|C{15mm}|C{16mm}|C{15mm}|C{16mm}}
			\hline
			
			\multirow{2}{*}{\textbf{Model}} &  \multicolumn{2}{|c|}{\texttt{\textbf{CIFAR}}$\mathbf{10}$} & \multicolumn{2}{|c}{\texttt{\textbf{CIFAR}}$\mathbf{100}$} \\
			\cline{2-5}
			\cline{2-5}
			& $64 \times 64$ & $128 \times 128$ & $64 \times 64$ & $128 \times 128$ \\
			\hline
			\hline
			\cite{dosovitskiy2020} & $93.75$ & $73.18$ & $69.53$ & $47.4$ \\
			\hline
			\cite{zhu2021long} & $93.58$ & $83.27$ & $74.47$ & $57.11$ \\
			\hline
			\cite{liu2022swin} & $\mathbf{97.66}$ & $84.9$ & $79.95$ & $58.59$ \\
			\hline
			\cite{didolkartemporal} & $94.79$ & $84.38$ & $79.17$ & $59.19$ \\
			\hline
			\textbf{BLRP} & $95.04$ & $\mathbf{86.85}$ & $\mathbf{83.85}$ & $\mathbf{61.06}$ \\
			\hline
	\end{tabular}
	}
	\caption{\footnotesize{\textbf{\textit{(Left)} LRA Results.} Our proposed approach surpasses state-of-the-art methods on all $3$ tasks from LRA benchmark. \textbf{\textit{(Right)} CIFAR Results.} The models were trained on the $64 \times 64$ setting and transferred to $128 \times 128$. Results are averaged across $5$ random seeds. \textbf{BLRP} outperforms the comparing baselines for all setups.}}
	\label{tbl:lra_cifar_results}
\end{table}

\begin{figure}[t]
	\begin{center}
	\scalebox{0.99}{
	\begin{tabular}{cc}
        \includegraphics[width=0.4\linewidth]{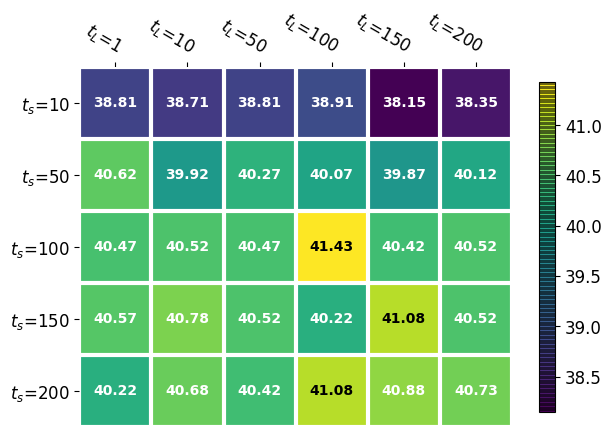} &
		\includegraphics[width=0.45\linewidth]{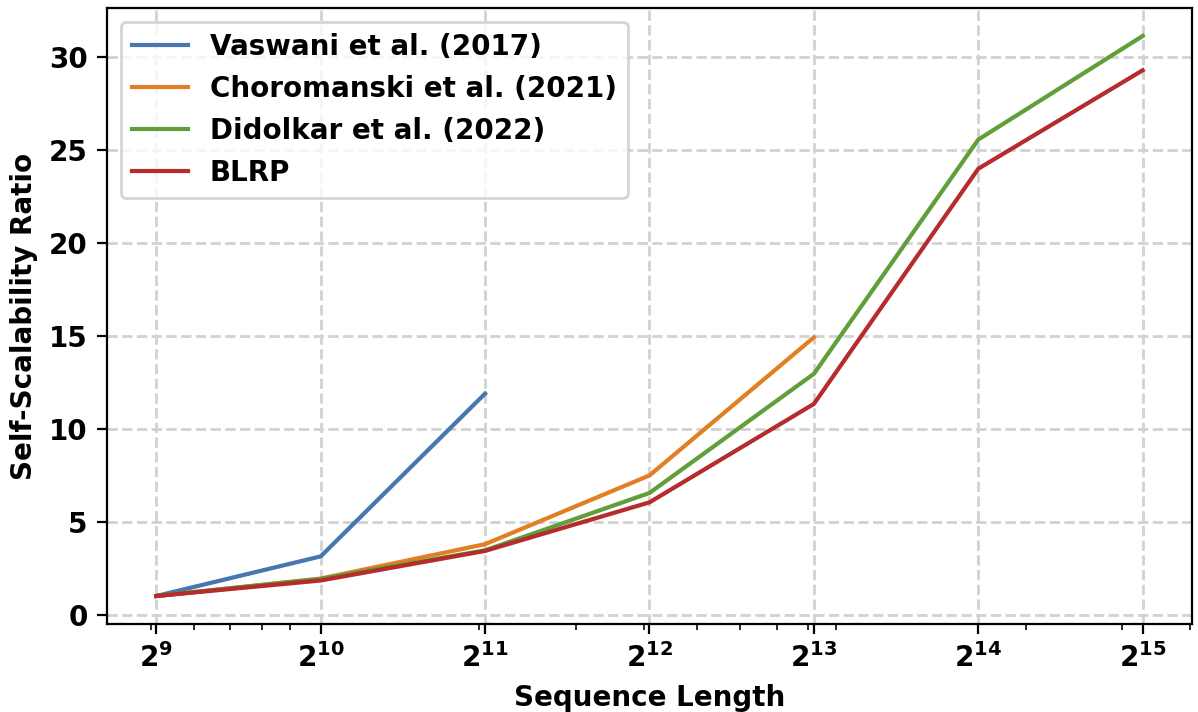}
	\end{tabular}}
	\end{center}
	\caption{\footnotesize{\textbf{\textit{(Left)} Impact of latent block size versus segment size.}  The rows correspond to segment sizes and are denoted with $t_S$ and the columns correspond to latent block sizes and are denoted with $t_L$. The highest performance (\textit{i.e.} $41.43$) is obtained with segment size and latent block sizes equal to the value of $100$. The poorest performance is obtained with a segment size of $1$ as the window context is very limited, thus the model not being able to infer the global information within the input sequence. Increasing the latent size is not enough to reach optimal performance. The local context has to be large enough to capture meaningful correlations. \textbf{\textit{(Right)} Scalability Analysis in Terms of GPU Memory Usage.} We tested the self-scalability in terms of GPU memory consumption for different sequence lengths (\textit{i.e.} $512$, $1024$, etc). For Transformer \cite{vaswani_nips_2017} and Performer \cite{choromanskirethinking} we were unable to test on extremely long sequences due to GPU memory limitations. All measurements are realised on an NVidia A$10$G machine with $24$ GB of memory.}}
	\label{fig:scalable_performance}
\end{figure}


\section{Experiments}
We demonstrate the effectiveness of \textbf{BLRP} on multiple modalities - textual modality by using  LRA \cite{taylong} and visual modality by using the CIFAR \cite{krizhevsky2009learning} benchmark. For all of our experiments we used the PyTorch \cite{NEURIPS2019_9015} library on a system with an Intel Xeon QuadCore $2.5$ GHz with $32$ GB system RAM and a single NVidia Tesla A$10$G GPU card with $24$ GB of GPU RAM. For training our proposed ensemble, we use AdamW optimiser \cite{LoshchilovH19} with a starting learning rate of $4e^{-4}$, $\beta_1 = 0.9$, $\beta_2 = 0.98$, $\epsilon = 1e^{-9}$, $\gamma = 0.8$ with linear learning rate decay. A detailed view on hyperparameters and architectural design choices for each task is available in Table \ref{tbl:reproducibility}.

\begin{table*}[t]
\begin{centering}
	\scalebox{0.7}{
		\begin{tabular}{C{32mm}|C{30mm}|C{26mm}|C{26mm}|C{26mm}|C{26mm}}
			\hline
			
			\textbf{Benchmark} & \textbf{Embedding Size} & \textbf{Hidden Size} & \textbf{$\#$ of Heads} & \textbf{Segment Size} & \textbf{Batch Size} \\
			\hline
			\hline
			\texttt{ListOps} & $64$ & $128$ & $8$ & $100$ & $32$ \\
			\hline
			\texttt{Text} & $256$ & $256$ & $4$ & $10$ & $24$ \\
			\hline
			\texttt{Retrieval} & $256$ & $736$ & $4$ & $100$ & $24$ \\
			\hline
			\texttt{CIFAR10} & $368$ & $736$ & $6$ & $10$ & $128$ \\
			\hline
			\texttt{CIFAR100} & $368$ & $736$ & $6$ & $10$ & $64$ \\
			\hline
	\end{tabular}}
	\caption{\footnotesize{\textbf{Hyperparameter details for BLRP on all benchmark.} This helps with reproducibility of our experimental results and provides the reader with meaningful insights on the internal settings of our best models.}}
	\label{tbl:reproducibility}
\end{centering}
\end{table*}

%


LRA consists of three major subtasks, \textbf{\texttt{ListOps}}, \textbf{\texttt{Text}} and \textbf{\texttt{Retrieval}} addressing classification objectives which encompass similarity, long-range dependencies and structural representation. It involves sequential data with input sequences ranging from $500$ up to $4,000$ elements. We use the recommended evaluation protocol described in \cite{taylong}. Moreover, our model has a total of $258K$ parameters which is in line with the requirements from \cite{taylong} for a fair comparison. 
In Table \ref{tbl:lra_cifar_results} \textit{(Left)} we showcase comparison results in terms of accuracy for all LRA tasks. On all $3$ tasks we achieve superior performance. The reported results are averaged over $4$ random runs. Similar to \cite{didolkartemporal}, for all experiments we use two self-attention layers, one cross-attention layer to update the segment embedding and one cross-attention layer to update the latent block states. 
For CIFAR we evaluate on the image classification task using \texttt{\textbf{CIFAR}}$\mathbf{10}$ and \texttt{\textbf{CIFAR}}$\mathbf{100}$ datasets. The models are trained using ViT \cite{dosovitskiy2020} backbone on a resolution of $64 \times 64$. The input image is split into patches of size $4 \times 4$ and fed in raster order to the model. Performance results are available in Table \ref{tbl:lra_cifar_results} \textit{(Right)}. The model is evaluated on the default $64 \times 64$ as well as the $128 \times 128$ configurations to stress the generalization capabilities for long sequences. \textbf{BLRP} outperforms both comparing baselines demonstrating the effectiveness of our proposed approach. Furthermore, with this experiment we demonstrate the versatility aspect of our attention-mechanism as it can be adapted to other backbone type using a different learning modality.





\subsection{Ablation Studies}
To better understand the limitations and the intuition behind our approach, we performed extensive ablation studies. In Figure \ref{fig:plot_performance} we plot the average performance of our model against \cite{didolkartemporal} for different sequence sizes for \texttt{ListOps} benchmark. It is noticeable that we achieve consistent superior performance for all ranges of sequence lengths (standard sequence lengths - leftmost part of the figure and extemely long sequences - rightmost part of the figure).
In Table \ref{tbl:ablation_bidirectional} we study the impact of the bidirectional sequence parsing heuristic as well as different alternatives for the usage of the latent block $\mathbf{L}$. We trained different variants of \textbf{BLRP} on \texttt{ListOps} benchmark by going on a unidirectional / bidirectional flow, and with various initializations of the latent block $\mathbf{L}$. We use $3$ different alternatives for $\mathbf{L}^\mathtt{INIT}$: \emph{(a)} $1$D positional embedding of segment elements denoted with $\mathtt{[1DPosEmb]}$, \emph{(b)} $\Theta(\mathbf{X})^\mathtt{INIT}$ corresponding to the learned dynamic projection of input sequence $\mathbf{X}$ which is added only at initialization of each pass (\textit{i.e.} forward or backward) and \emph{(c)} $\Theta(\mathbf{X})$ corresponding to the learned dynamic projection added throughout the sequence processing as a residual connection and within the cross-attention operation. This process is applied in a unidirectional fashion, going forward, rows $2-4$, going backward, rows $5-7$ and bidirectionally, rows $8-14$, where we use different initialisations for the latent block of each pass. The best performance is achieved while using the dynamic projection function for both passes and adding the initialization as a skip connection throughout the process. This validates the importance of the bidirectional flow combined with information gain brought by using the dynamic projection.

\begin{wraptable}{r}{6.6cm}
    \scalebox{0.6}{\begin{tabular}{C{19mm}|C{29mm}|C{33mm}|C{13mm}}
			\hline
			
			\textbf{Direction} &  \textbf{Forward} $\mathbf{L}$ \textbf{Usage} & \textbf{Backward} $\mathbf{L}$ \textbf{Usage} & $\mathbf{Acc.}$ \\
			\hline
			\hline
			\multirow{3}{*}{Forward} & $\mathtt{[1DPosEmb]}$ & \multirow{3}{*}{$N/A$} & $39.84$ \\
			& $\Theta(\mathbf{X})^\mathtt{INIT}$ & & $39.91$ \\
			
			& $\Theta(\mathbf{X})$ & & $37.00$ \\
			\hline
			\multirow{3}{*}{Backward} & \multirow{3}{*}{$N/A$} & $\mathtt{[1DPosEmb]}$ & $40.38$  \\
			& & $\Theta(\mathbf{X})^\mathtt{INIT}$ & $40.47$ \\
			
			& & $\Theta(\mathbf{X})$ & $40.94$ \\
			\hline
			\multirow{7}{*}{Bidirectional} & $\mathtt{[1DPosEmb]}$ & $\mathtt{[1DPosEmb]}$ & $39.44$\\
			& $\mathtt{[1DPosEmb]}$ & $\Theta(\mathbf{X})^\mathtt{INIT}$ & $39.44$ \\
			
			& $\Theta(\mathbf{X})^\mathtt{INIT}$ & $\mathtt{[1DPosEmb]}$ & $39.55$\\
			& $\Theta(\mathbf{X})^\mathtt{INIT}$ & $\Theta(\mathbf{X})^\mathtt{INIT}$ & $39.40$\\
			& $\mathtt{[1DPosEmb]}$ & $\Theta(\mathbf{X})$ & $40.86$ \\
			
			& $\Theta(\mathbf{X})$ & $\mathtt{[1DPosEmb]}$ & $40.66$\\
			& $\Theta(\mathbf{X})$ & $\Theta(\mathbf{X})$ & $\mathbf{41.43}$\\
			\hline
\end{tabular}}
\caption{\footnotesize{\textbf{Importance of bidirectional flow.} To validate the impact of the bidirectional flow, we trained different variants of \textbf{BLRP} on \texttt{ListOps} benchmark by going on a unidirectional / bidirectional flow, and with various initializations of the latent block $\mathbf{L}$. The best performance is achieved for a bidirectional flow, using dynamic projection initialization combined with skip connections.}}
\vspace{-5mm}
\end{wraptable}
\label{tbl:ablation_bidirectional}

In Figure \ref{fig:scalable_performance} \textit{(Left)} we analysed the impact of the latent block size against the segment size on the \texttt{ListOps} task.  We trained and evaluated the performance of the model on all the combinations of latent block and segment sizes from the following set of values $[1, 10, 50, 100, 150, 200]$ and $[10, 50, 100, 150, 200]$, respectively. The rows correspond to segment sizes and are denoted with $t_S$ and the columns correspond to latent block sizes and are denoted with $t_L$.  We notice a performance drop correlated with the decrease of segment size. The highest performance (\textit{i.e.} $41.43$) is obtained with segment size and latent block sizes equal to the value of $100$. The worst performance is obtained with a segment size of $10$ as the window context is very limited, thus the model not being able to capture  powerful local correlations that would enable learning discriminative global representations. In essence, a limitation of the context window leads to poorer understanding of the entire sequence. Moreover, we observe that the performance is higher when the latent size is equal to the segment size. Another important study is emphasized in Table \ref{tbl:ablation_theta}. We use different alternatives for building $\mathbf{L}^\mathtt{INIT}$ with $\Theta$ function. Firstly, we use the same variable generated by $\Theta$ function to instantiate the forward and backward pass. Secondly, we use the $\Theta$ function as a Siamese component to instantiate $2$ separate variables for the bidirectional pass. Lastly, we learn $2$ separate representations of the $\Theta$ function, one for the forward and one for the backward pass. The best outcome is obtained with different initialisations and different learned dynamic projection heads. This is a result of the fact that we strengthen the signal on both sides of the sequence pass using global information synthesised from the entire sequence.

To consolidate the long-range context understanding claim, we performed a sequence length augmentation study on \texttt{ListOps} task in table \ref{tbl:long-range-analysys}. We artificially multiplied the original sequences by self-concatenating them via the \texttt{MAX} operator. Thus, we ensure the output is invariant w.r.t. the sequence length. This process was achieved with a multiplication factor of between $1$ and $4$, thus forcing the model to process sequences of up to $8192$ tokens. For testing, we use the models trained on the original \texttt{ListOps} containing sequences of up to $2048$ elements. Our approach is able to achieve high performance, although it is tested out-of-domain sequences in terms of length. Comparatively, the model of \cite{didolkartemporal} manifests a high performance drop for all sequence lengths. This observation is in line with the rightmost section of the plot from Figure \ref{fig:plot_performance}, showcasing the capability of our approach to maintain a long-context understanding despite the high length of the processed sequence.

\begin{table*}[t]
\begin{centering}
	\scalebox{0.75}{
		\begin{tabular}{C{140mm}|C{50mm}}
			\hline
			
			\textbf{Latent Variant} & $\mathbf{Accuracy}$ \\
			\hline
			\hline
			Identical $\mathbf{L}^\mathtt{INIT}$ for backward / forward pass & $40.86$ \\
			\hline
			Separate $\mathbf{L}^\mathtt{INIT}$ obtained with Siamese $\Theta(\mathbf{X})$ for backward / forward pass & $41.32$ \\
			\hline
			Separate $\mathbf{L}^\mathtt{INIT}$ obtained with separate $\Theta(\mathbf{X})$ for backward / forward pass & $\mathbf{41.43}$ \\
			\hline
	\end{tabular}}
	\caption{\footnotesize{\textbf{Ablation on usage of dynamic projection function $\Theta$.} We experimented using different alternatives for $\Theta$ function for the bidirectional setup. The best results are obtained when initializing the forward and backward pass latent blocks using separately learned dynamic projection functions. The second best is when we have different representations obtained with a Siamese setup of $\Theta$.}}
	\label{tbl:ablation_theta}
\end{centering}
\end{table*}








\begin{wraptable}{r}{8cm}
	\scalebox{0.75}{
		\begin{tabular}{C{35mm}|C{11mm}|C{11mm}|C{11mm}|C{11mm}}
			\hline
			\multirow{2}{*}{\textbf{Model}} &  \multicolumn{4}{|c}{\textbf{Sequence Multiplication Factor}}  \\
			\cline{2-5}
			& $\times 1$ & $\times 2$ & $\times 3$ & $\times 4$ \\
			\hline
			\hline
		    \cite{didolkartemporal} & $38.2$ & $16.98$ & $15.70$ & $14.86$ \\
			\hline 
			\textbf{BLRP} & $41.43$ & $36.04$ & $35.84$ & $35.79$ \\
			\hline
	\end{tabular}}
	\caption{\footnotesize{\textbf{Performance analysis over long-range context.} We report the average performance on artificially augmented sequences from \texttt{ListOps}. The augmentation is performed by self-concatenating input sequences via the \texttt{MAX} operator, thus ensuring we obtain the same output. Our approach is able to maintain a high performance threshold, whereas \cite{didolkartemporal} gets a large performance gap due to the domain-shift distribution caused by sequence length.}}
	\label{tbl:long-range-analysys}
	\vspace{-4mm}
\end{wraptable}

\subsection{Computational Efficiency and Limitations}
Our proposed approach is incomplete without a study in terms of computational efficiency and scalability. To demonstrate the capabilities for our proposed \textbf{BLRP} with respect to this demand, we conducted an experiment in terms of self-scalability of the GPU memory consumption correlated with an increase in terms of sequence size. We measured how the model scales in terms of memory consumption, as the input sequences are increased considerably in length (\textit{i.e.} exponential increase as a power of $2$).  Results are illustrated in Figure \ref{fig:scalable_performance} \textit{(Right)}. Basically, this study highlights how the model scales with respect to the length of the sequential data being processed.  Please note that the vanilla transformer \cite{vaswani_nips_2017} is limited to sequences of up to $4096$ elements due to GPU memory limitations. It is worth noticing that our scalability is directly proportional against TLB \cite{didolkartemporal}, achieving almost linear scalability, however at a higher performance gain (see Figure \ref{fig:plot_performance}). 
In terms of inference speed measurements, we have averaged the run times normalized per sequence length and compared against \cite{didolkartemporal} baseline ($0.39582$ seconds for TLB \cite{didolkartemporal} compared with $0.41743$ seconds for our method, both methods measured with a batch with $32$ elements with $10,000$ elements) demonstrating that we achieve similar time efficiency for extremely long sequences at an overall higher performance gain (see Figure \ref{fig:plot_performance}).

As shown in our ablation studies, the proposed method is extremely sensitive w.r.t. the right combination of latent block size and segment size. Thus, making this type of methodology unsuitable for adapting pre-trained models on new data domains. For extremely long sequences we do not have an explicit gating mechanism, thus we cannot provide an insurance that forgetting irrelevant information will naturally occur via the cross-attention operations. This might explain why increasing the segment size is correlated with increasing the latent size. Moreover, we do not have a natural flow of combining multi-modal information flows.

\section{Conclusion}

With our proposed \textbf{BLRP} framework we demonstrate the importance of bidirectionally modelling close and distant dependencies across long sequences. We prove the optimality with respect to the long-sequence parsing aspect of \textbf{BLRP} on challenging benchmarks from multiple domains (visual and textual) using different learnable backbones. Furthermore, through our ablation studies, we emphasize the innovative aspects, the limitations and the intuition behind our work and the individual contribution of all architectural design choices.



\bibliography{main}

\end{document}